\documentclass{article}
\usepackage{ijcai17}

\usepackage{times}

\usepackage{latexsym}
\usepackage{graphicx}
\usepackage{subfigure}
\usepackage{flushend}
\usepackage{url}
\usepackage{color}
\usepackage[small]{caption}

\title{Inverted Bilingual Topic Models for Lexicon Extraction from Non-parallel Data}
\author{Tengfei Ma\\ 
IBM T. J. Watson Research Center  \\
Tengfei.Ma1@ibm.com
\And
Tetsuya Nasukawa\\ 
IBM Research-Tokyo \\
NASUKAWA@jp.ibm.com}

\begin{document}

\maketitle

\begin{abstract}
% A good lexicon is an important resource for various cross-lingual tasks such as information retrieval and text mining. In this paper, we focus on extracting translation pairs from non-parallel cross-lingual corpora. Previous lexicon extraction algorithms for non-parallel data generally rely on an accurate seed dictionary and extract translation pairs by using context similarity. However, there are two problems. One, a lot of semantic information is lost if we just use seed dictionary words to construct context vectors and obtain the context similarity. Two, in practice, we may not have a clean seed dictionary. For example, if we use a generic dictionary as a seed dictionary in a special domain, it might be very noisy. To solve these two problems, we propose two new bilingual topic models to better capture the semantic information of each word while discriminating the multiple translations in a noisy seed dictionary. We then use an effective measure to evaluate the similarity of words in different languages and select the optimal translation pairs. Results of experiments using real Japanese-English data demonstrate the effectiveness of our models.
Topic models have been successfully applied in lexicon extraction. However, most previous methods are limited to document-aligned data. In this paper, we try to address two challenges of applying topic models to lexicon extraction in non-parallel data: 1) hard to model the word relationship and 2) noisy seed dictionary. To solve these two challenges, we propose two new bilingual topic models to better capture the semantic information of each word while discriminating the multiple translations in a noisy seed dictionary. We extend the scope of topic models by inverting the roles of "word" and "document". In addition, to solve the problem of noise in seed dictionary, we incorporate the probability of translation selection in our models. Moreover, we also propose an effective measure to evaluate the similarity of words in different languages and select the optimal translation pairs. Experimental results using real world data demonstrate the utility and efficacy of the proposed models.
\end{abstract}

\section{Introduction}
\label{intro}
Bilingual lexicons play an important role in cross-lingual information retrieval and text mining tasks. However, there is often no existing dictionary for some technical data or low-resourced language pairs. For example, in some special domains, there are always novel words or new expressions emerging, and a generic dictionary can hardly keep up with them. Thus, automatically extracting translation pairs \cite{andrade2010robust},\cite{bollegala2015cross} has attracted a lot of attention. 

As one of the most successful methods for latent semantic analysis in the past years, topic models (e.g. Latent Dirichlet Allocation \cite{blei2003latent} ) have demonstrated usefulness for lexicon extraction \cite{vulic2011identifying,vulic2013cross,mimno2009polylingual,ni2009mining}. 
% A classical bilingual LDA generally requires the documents to be aligned in pairs. The basic idea is that an aligned document pair should have the same topic distribution $\theta$. For each document pair $<d_{l1},d_{l2}>$, a topic distribution $\theta$ is drawn from a Dirichlet distribution: $\theta \sim \rm{Dirichlet}(\alpha)$. Then, for each language $l$, a topic assignment is sampled for each word, as $z^l \sim \rm{Multinomial}(\theta)$. As the final step, words in each language are separately drawn from their topic assignment and topic-specific distribution $\phi^l_{z^l} \sim \rm{Dir}(\beta^l)$: $w^l \sim \rm{Multinomial}(\phi^l_{z^l})$.
% In this way, the topics in different languages can be connected. Moreover, we can measure the similarity of documents in different languages, $d_i, d_k$, simply by computing the similarity of their topic distributions $Sim(\theta_i, \theta_k)$. 
A classical bilingual topic model utilizes the alignment relationship of documents in different languages, and projects them into a shared latent semantic space. The basic idea is that an aligned document pair should have the same topic distribution, thus the topics in different languages can be connected. Then the shared topics could be used to calculate the similarity of words in different languages, e.g. through Bayesian rules and K-L divergence \cite{vulic2011identifying}. Despite their success, there are two problems of the bilingual topic models in lexicon extraction. 

Firstly, almost all previous cross-lingual topic models are developed for document-aligned data, however in practice corpora are often not aligned.  Extracting lexicons from non-parallel corpora is more valuable yet far more challenging. Although topic models could easily integrate the document relationship \cite{dietz2007unsupervised}, \cite{chang2009relational}, they have difficulties representing the word relationship, which is needed for lexicon extraction from non-parallel data. 

Secondly, to our best knowledge, few of existing bilingual topic models considered the probability of the multiple translations in their models, or handled the noise issue in their dictionary. Although some initial effort has been made, for example, Boyd-Graber and Blei \shortcite{boyd2009multilingual} integrated the prior of word matching to the bilingual topic models in non-parallel data. However, their results indicate the model had no effect of finding new word translations.

In this work, we propose two novel bilingual topic models to extend their applications to lexicon extraction from non-parallel data. We reverse the roles of documents and words to represent each word as a pseudo document and then model the words rather than original documents. To be specific, we use inverted indexing to represent a word as a list of documents where it occurs. After obtaining the pseudo documents, each word is then modeled as a topic distribution. Different from some previous work related to cross-lingual inverted indexing \cite{sogaard2015inverted}, we do not consider connections between documents but only between words. Each translation pair is assumed to own the same topic distribution. In this way, topics in different languages can also be connected.

%In this paper, we propose two new bilingual topic models to extend the application of topic models to lexicon extraction in non-parallel data. Considering that topic models could easily integrate the document relationships, 
%we develop a new approach to topic modeling by reversing the roles of documents and words in a topic model. We first represent each word as a pseudo document and then model the words instead of the original documents. Concretely, in this work we use inverted indexing to represent a word as a list of documents in which it occurs. After obtaining the pseudo documents, we use our topic models to model each word as a topic distribution. Different from the motivation of previous work related to cross-lingual inverted indexing \cite{sogaard2015inverted}, we do not consider connections between documents but only between words. Each translation pair is assumed to own the same topic distribution. In this way, the topics in different languages can also be connected. 

Next, in order to solve the problem of noisy translations in the seed dictionary, in the generative process of our models we select only one proper translation from all candidates in the seed dictionary probabilistically instead of using all of them equivalently. 
In addition, in the original bilingual LDA\cite{vulic2011identifying} only the aligned document pairs are modeled. In our case, however, we only have a subset of words included in the seed dictionary. The remaining words do not have any connection with words in other languages, but they are also included in our models, making the models essentially semi-supervised.
We use Gibbs sampling for posterior inference and get the similarity between words across languages on the basis of their topic distributions. In contrast to traditional cosine similarity and KL divergence, we define the similarity measure as the probability of a word generating another. Given a word in a source language, the word with the most similar topic distributions in the target language is then regarded as its translation. 

To summarize, it is worthwhile to highlight the following contributions of the proposed model:
\begin{itemize}
\item We adopt the inverted indexing technique to extend the scope of topic models to the task of lexicon extraction from non-parallel data.
\item We extend the classical bilingual LDA by incorporating the probability of multiple translations in the model, thus solving the noise issue for the seed dictionary.
\item We propose a new similarity measure from the conditional generating probability for two words across languages to handle the correlation of topics. Experimental results demonstrate its advantage over other traditional measures.
\end{itemize}

The following sections are organized as follows: we first review related work in Section \ref{sec:related}. Then we propose our problem definition and new models in Section \ref{sec:model}. Section \ref{sec:measure} explains how we use our topic model to measure the word similarity. In Section \ref{sec:experiment} we describe our experiments. At last we conclude our work in Section \ref{sec:conclusion}.

\section{Related Work}
\label{sec:related}
We have introduced the topic modeling methods for lexicon extraction. In this section, we will focus more on the history of {\bf lexicon extraction from non-parallel data}.

The most well-established work on lexicon extraction is based on word alignment in parallel datasets. We can easily use a statistical machine translation system \cite{lopez2008statistical} to induce translation pairs from parallel data. However, parallel data is not plentiful for all language pairs or all domains. This restricts the usefulness of these methods. 

Lexicon extraction from non-parallel data was pioneered by \cite{rapp1995identifying} and \cite{fung1998ir}. Instead of parallel/comparable documents, they use a seed dictionary as the pivots. Generally, this kind of approach can be factorized into two steps: 1, construct a context vector for each word, and 2, compute the context similarities on the basis of pivot words (i.e., seed dictionary entries). A common hypothesis is that a word and its translation tend to occur in similar contexts. Previous research has defined various correlation measures to construct a context vector representation for a word, including tf-idf \cite{fung1998ir} and pointwise mutual information (PMI) \cite{andrade2010robust}. As for the similarity computation, cosine similarity \cite{fung1998ir}, non-aligned signatures (NAS) \cite{shezaf2010bilingual}, and Johnson-Shannon divergence \cite{pekar2006finding}, etc. can be used.

The context similarity-based models rely on the quality and the size of seed dictionaries. When a seed dictionary is small, the context vector will be too sparse and the similarity measure is not accurate enough. Recent work has used graph-based methods to propagate the seed dictionaries \cite{laws2010linguistically,tamura2012bilingual}. There are also some methods that project the word vectors in different languages into the same low-dimensional space, such as linear transformation for cross-lingual word embedding \cite{mikolov2013exploiting}. Our motivation is similar to the graph-based and word embedding-based models in that we use a topic model to represent each word as a topic distribution in order to avoid the sparseness of context vectors. However, while the previous approaches generally just select the reliable translations as seeds \cite{mikolov2013exploiting,haghighi2008learning}, we assume our seed dictionary is noisy. We add the probability of existing translations as a new latent variable and make our model more robust and generalizable. Most recently, Duong et. al.\shortcite{duong2016learning} and Zhang et. al. \shortcite{zhang2017bilingual} propose new bilingual word embedding methods to deal with the noise of the seed dictionary. But the word embedding methods cannot explicitly interpret the uncertainty of the multiple translations. As generative models, our topic models have obviously better interpretability for translation selection.

\section{Proposed Model}
\label{sec:model}
In this section, we start by making a formal definition of our problem. Then we will describe the details of our new model, following a brief introduction to the background knowledge of cross-lingual topic models. 
\subsection{Problem Definition}
\label{sec:def}

Assume that we are given only two mono-lingual corpora in different languages, $C^e$ and $C^j$. They are neither sentence-aligned nor document-aligned, but are in the same domain. The documents in $C^e$ are noted as $\{d^e_i\}$ for $i = 1,...N^e$ where $N^e$ is the number of documents in $C^e$; while the documents in $C^j$ are noted as $\{d^j_i\}$ for $i = 1,...N^j$ where $N^j$ is the number of documents in $C^j$. Other than the data corpora, we also have a set of seed dictionaries. We assume that the seed dictionary comes from the generic domain, and is noisy. It means one term in the seed dictionary may have several translations, within which some translations are not correct in this domain.

Now given a term in the source language $t^j$ which appears in $C^j$, we want to find the most possible translation term in $C^e$.

\subsection{Model Description}
\subsubsection{Inverted Indexing and Topic Models}
\label{sec:BiLDA} 
Our approach to lexicon extraction is to first use topic models to model the cross-lingual data and obtain the topic distribution of each word. Then we can compare the topic distributions to compute the word similarities and get the translation. 

In a conventional topic model, only the documents are represented by topic distributions, while the topic distribution for a word is not explicit. In addition, it is relatively easy to model document pairs or other document relationships by various topic models, as discussed earlier. However, in our problem setting, we only have a seed dictionary and non-parallel data corpora, so it is difficult to find document relationships but easy to get word translation pairs. The motivation is that if we can transfer a word into a pseudo document, we can utilize the word relationship in seed dictionaries.

In order to implement this idea, we invert the document-word index so that a word is constructed by a list of document IDs. If we assume a word $w$ that appears in $d_1$ twice, $d_2$ once, and $d_3$ once, it is represented as $(d_1, d_1, d_2, d_3)$. We also keep the word frequency in this representation.

As far as we know, this is the first work to integrate inverted indexing and topic models. We can also use other ways to construct the pseudo documents, such as using neighbor words. However, there are far fewer documents than context words, so we can reduce the computational cost. In addition, using inverted indexing-based representation enables us to easily calculate $p(d|w) = \sum_z p(d|z)p(z|w)$ from the topic distributions. So we can easily achieve the conditional probability of all documents when given a search term in another language. This might be useful for cross-lingual information retrieval tasks (although this is not our focus in this paper).

To avoid confusion, in the following sections we use "word'' to indicate the pseudo document in topic models and use ``document'' to indicate the basic element in a pseudo document. Thus, a topic is a distribution of documents, and a word is a mixture of topics. That is to say, we have reversed the roles of words and documents in conventional topic models.

Once we obtain the pseudo documents, we can use them to train a Bilingual LDA model\cite{vulic2011identifying}. If two words are translations to each other, they are assumed to have similar topic distributions. The problem is that we only have a subset of words that are translated, and a word in a seed dictionary may have several translations. Therefore, first we need to construct one-to-one word pairs, the same as what Bilingual LDA does for documents.

Intuitively, it is not a good choice to make all translations modeled because the different translations will own the same topic distribution if a word has polysemy. Instead, we just select the most frequent term in the translation list to form a translation pair. Then, for all translation pairs, we use the same model as the Bilingual LDA. Words that do not have translations are modeled together using the original LDA.

\begin{itemize}
\item For each translation pair {$t^j$, $t^e$}, 
\begin{itemize}
\item Sample a topic distribution $\theta \sim \rm{Dirichlet}(\alpha)$
\end{itemize}
\item For each word $t_l$ ($l \in \{j,e\}$) without translation, 
\begin{itemize}
\item Sample a topic distribution $\theta_l \sim \rm{Dirichlet}(\alpha)$
\end{itemize}
\end{itemize}

Following this process, we sample the topics for each token $d^e$ and $d^j$ from $\theta$ and then draw documents from the topic. We also performed experiments to try out another way to obtain translation pairs. Instead of just selecting one translation, we randomly select a translation in each sample iteration, which means we finally use all the translations over all iterations. We call this model {\bf BiLDA\_all}, while the previous one is called {\bf BiLDA}. A comparison of the two models is given in Section \ref{sec:experiment}. They are both used as our baseline systems. 

If we select just one translation, there is a risk of losing a lot of information. This is especially problematic when the seed dictionary is not large, as the lost information will cause a serious performance decrease. On the other hand, using all translations without discrimination is not ideal either, as we discussed previously. We therefore came up with a solution to properly select the correct translation for each word. 

\subsubsection{ProbBiLDA}
We developed two approaches to model the probability of translation selection. The first approach is to add a selection variable for each token (i.e. each document) $d^j$ in word $t^j$, such that the topic distribution of each $t^j$ is a mixture of its translations. This is similar to the idea of citation models \cite{dietz2007unsupervised}, which model the probability of citation as the influence rate. The difference is that we have two sets of topics for the two respective languages. We do not directly share the topics of the "cited'' pseudo document, opting instead to use the ``cited'' topic distribution to sample a new topic in its own language. We call this model {\bf ProbBiLDA} (probabilistically linked bilingual LDA). The generative process of the target language $e$ is same as the original LDA. To save space, we only list the generative process for source language $j$ as follows.

\begin{itemize}
\item For each topic $z^l \in \{1,...K\}$ (language $l \in \{j,e\}$):
\begin{itemize}
\item Sample a document distribution $\phi^l_{z^l} \sim \rm{Dir}(\beta)$
\end{itemize}

\item For each word $t^{j}$:
\begin{itemize}
\item If $t^j$ not in seed dictionary::
\begin{itemize}
\item Sample a topic distribution $\theta_{t^j} \sim \rm{Dir}(\alpha)$
\item For each position $i$ in the word: 
\begin{itemize}
\item Sample a topic assignment $z^j_i \sim \rm{Multi}(\theta_{t^j})$
\end{itemize}
\end{itemize}
\item If $t^j$ has $S$ translations:
\begin{itemize}
\item Draw a probability distribution $\psi_{t^j} \sim \rm{Dir}(\alpha_\psi)$ over all translations
\item For each position $i$ in the word: 
\begin{itemize}
\item Sample a translation $s_{i} \sim \rm{Multi}(\psi_{t^j})$ from the $S$ translations in language $e$
\item Draw a topic $z^j_i \sim \rm{Multi}(\theta_{s_i})$ 
\item Draw a document $d^j_i \sim \rm{Multi}(\phi^j_{z^j_i})$
\end{itemize}
\end{itemize}
\end{itemize}
\end{itemize}

\subsubsection{BlockProbBiLDA}
Another way to model the probability of translations is to add the probability variable to the word itself instead of to each document in that word. That is to say, we select a translation for the whole word, and all the documents in that word follow the same topic distribution. For example, a word $t = (d1, d2)$ has three translations $t1, t2, t3$. If we use the ProbBiLDA, the topic of each document in word $t$ is sampled from different translations, e.g., $z_{d1} \sim \theta_{t2}$ and $z_{d2} \sim \theta_{t3}$. However, in the new model, we require that all documents in $t$ can only select one same translation in each iteration. If $t2$ is selected as the translation of $t$, we must have $z_{d1} \sim \theta_{t2}$ and $z_{d2} \sim \theta_{t2}$.

 \begin{figure}[t]
      \centering
      \subfigure[ProbBiLDA]{
      \includegraphics[width=0.17\textwidth]{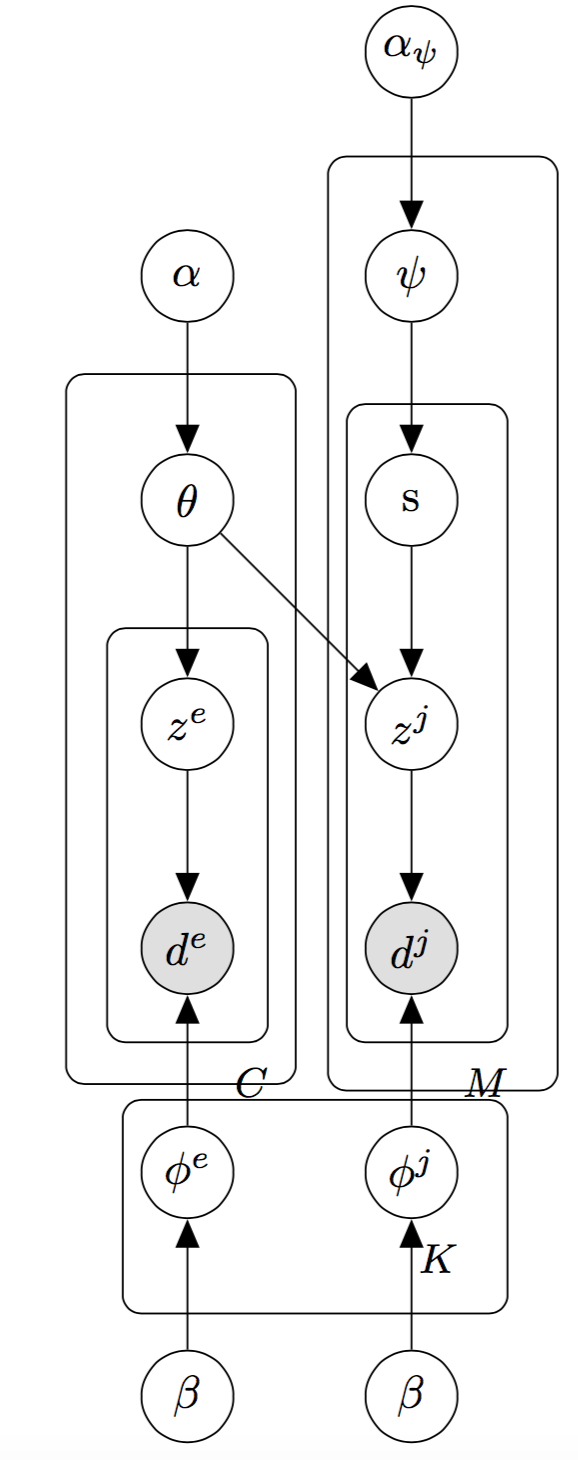}
      }
\hspace{0.05\textwidth}
      \subfigure[BlockProbBiLDA]{
      \includegraphics[width=0.17\textwidth]{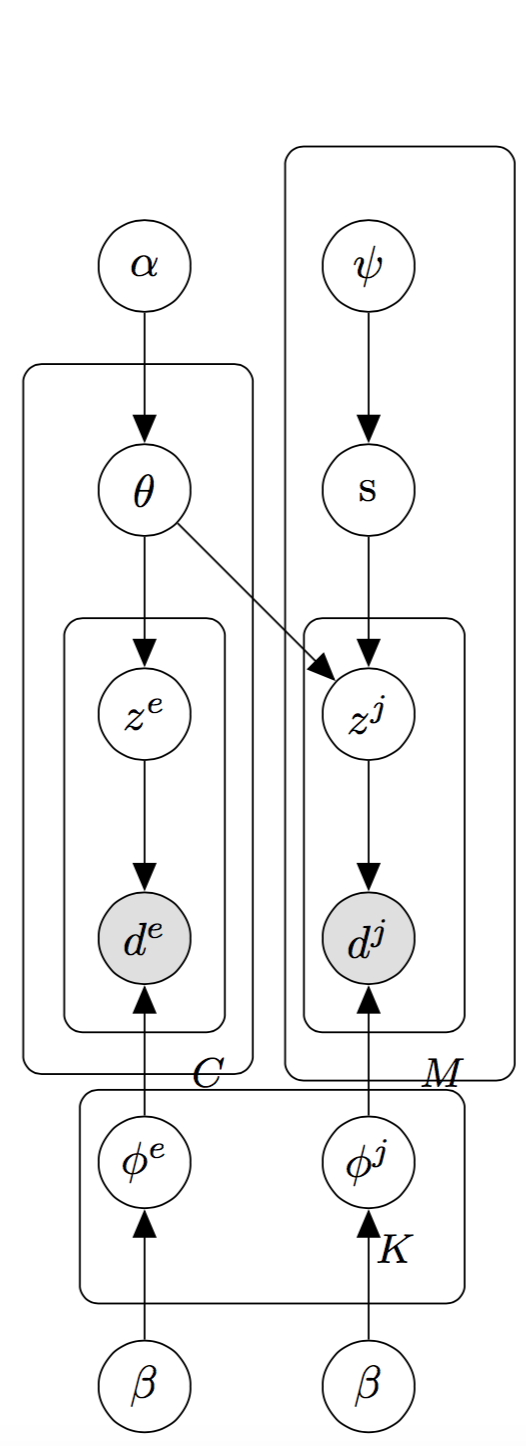}
      }
      \caption{Graphic Representation of the two new Bilingual topic models. Note that in the figure we only represent the generative process for words in seed dictionaries; while for other words, they are modeled as same as an original LDA.}
      \label{fig:model}
\end{figure}
As all the documents select translations together like a block, we call this model {\bf BlockProbBiLDA}. This model is essentially more similar to the original Bilingual LDA. Compared to Bilingual LDA, it does not fix the translation pairs but rather assigns a prior to each translation. Compared to the generative process of ProbBiLDA, it only changes the position of $s$ and uses a uniform prior distribution $\psi$ instead of Dirichlet prior. The graphical representation of BlockProbBiLDA is shown in Figure \ref{fig:model}; and its generative process for words in source language is as follows:

\begin{itemize}
\item For each word $t^j$:
\begin{itemize}
\item If $t^j$ not in seed dictionary:
\begin{itemize}
\item Sample a topic distribution $\theta_{t^j} \sim Dir(\alpha)$
\item For each position $i$ in the word:
\begin{itemize}
\item Sample a topic assignment $z^j_i \sim \rm{Multi}(\theta_{t^j})$
\end{itemize}
\end{itemize}
\item If $t^j$ has $S$ translations.
\begin{itemize}
\item Sample a translation $s \sim Multi(\psi_{t^j})$ from the $S$ translations, where $\psi_{t^j} $ is a uniform distribution over all translations
\item For each position $i$ in the word:
\begin{itemize}
\item Draw a topic  $z^j_i \sim Multi(\theta_{s})$, 
\item Draw a document $d^j_i \sim Multi(\phi^j_{z^j_i})$
\end{itemize}
\end{itemize}
\end{itemize}
\end{itemize}

\subsection{Posterior Inference}
For both of the two new models, we use collapsed Gibbs sampling to approximate the posterior. We iteratively update latent variables (including topic assignment $z$) given other variables. 
\subsubsection{Posterior Inference for ProbBiLDA}
\label{sec:infprob}
In the model of ProbBiLDA, for each document $d^j_{t^j,i}$ in a word $t^j$, we assume that it selects a translation word $c$ in target language $e$, i.e. it is drawn from the topic distribution of this word. Given the translation selection, and other topic assignments, we sample the topic for document $d^j_{t^j,i}$ according to:
\begin{eqnarray}
\label{sampling:topic}
&  p(z^j_i = k|z^j_{-i, t^j}, s_i = c, d^j_{t^j,i}=n , \theta) \\ \nonumber
& \propto \frac{nmk(c,k)+cmk(c,k)+\alpha -1}{nm(c) +cm(c) +K * \alpha -1} * \frac{ nkvj(k,n) + \beta -1} {nkj(k) + V_j*\beta -1}
\end{eqnarray}
where $nmk(c,k)$ denotes the number of documents in word $c$ that are assigned to  topic $k$; $nm(c)$ denotes the total number of documents in word $c$; $cmk(c,k)$ is the number of documents with topic $k$ in language $e$ that select $c$ as the translation of its associated word; and $cm(c)$ is the total number of documents in language $e$ with translation selection $c$. $nkvj(k,n)$ is the number of times when document $n$ is assigned to topic $k$ in language $j$; and accordingly $nkj(k)$ is the sum of $nkvj(k,n)$ over all documents in language $j$; $V_j$ is the total number of documents in language $j$.
 
Given these topic assignments, we can sample the translation selection:
\begin{eqnarray}
&  p(s_i = c|s_{-i}, z^j_i =k, d^j_{t^j,i}=n , \theta) \\ \nonumber
&\propto \prod_i \frac{nmk(c,k)+cmk(c,k)+\alpha -1}{nm(c) +cmk(c) +K * \alpha -1}  * \frac{nms(t^j, c) + \alpha_\psi -1}{nm(t^j) + S(t^j) * \alpha_\psi -1} 
\end{eqnarray}
where $nms(t^j, c)$ denotes the number of documents in word $t^j$ which selects translation $c$; and $S(t^j)$ is the number of translation candidates for word $t^j$.

The above sampling scheme is for the source language. While for target language, we only need to care about the topic assignments.
\begin{eqnarray}
\label{sampling:tar}
&  p(z^e_i = k|z^e_{-i, t^e}, d^e_{t^e,i}=n , \theta) \\ \nonumber
& \propto \frac{nmk(t^e,k)+cmk(t^e,k)+\alpha -1}{nm(t^e) +cm(t^e) +K * \alpha -1} * \frac{ nkve(k,n) + \beta -1} {nke(k) + V_e*\beta -1}
\end{eqnarray}
where the denotations of the variables are similar to the ones defined in (\ref{sampling:topic}).
 
Given all the topic assignments, we can then derive the topic distribution $\theta_{m}  = (\theta_{m,1}, \theta_{m,2},...,\theta_{m,K})$ for word $m$.
\begin{equation}
\theta_{m,k} = \frac{nmk(m,k)+\alpha}{nm(m) +K * \alpha} 
\end{equation}
The topic variables are derived from:
\begin{equation}
\phi^e_k = \frac{ nkve(k,n) + \beta } {nke(k) + V_e*\beta }
\end{equation}
\begin{equation}
\phi^j_k = \frac{ nkvj(k,n) + \beta } {nkj(k) + V_j*\beta }
\end{equation}
We run 1500 iterations for inference while the first 1000 iterations are discarded as burn-in steps. After the sampling chain converges, we average the value of $\theta_{m}$ to get the final per-word topic distribution.
 
\subsubsection{Posterior Inference for BlockProbBiLDA}
For each word $t^e$, we sample its topic according to:
\begin{eqnarray}
&  p(z^e_i = k|z^e_{-i, t^e}, d^e_{t^e,i}=n , \theta) \\ \nonumber
& \propto \frac{nmk(t^e,k)+cmk(t^e,k)+\alpha -1}{nm(t^e) +cm(t^e) +K * \alpha -1} * \frac{ nkve(k,n) + \beta -1} {nke(k) + V_e*\beta -1}
\end{eqnarray}

For each word $t^j$, if it is in the dictionary, and it selects $c$ as its translation in the previous iteration, then

\begin{eqnarray}
&  p(z^j_i = k|z^j_{-i, t^j}, d^j_{t^j,i} = d, \theta) \\ \nonumber
& \propto \frac{nmk(t^j, k) + nmk(c,k)+\alpha -1}{nm(t^j) + nm(c) +K * \alpha -1} * \frac{ nkvj(k,d) + \beta -1} {nkj(k) + V_j*\beta -1}
\end{eqnarray}

The selection of translations is sampled by
\begin{eqnarray}
&  p(s^j = t^e|z^j, d^j_{t^j,i} =d, \theta) \\ \nonumber
&\propto \prod_i \frac{nmk(t^e, z^j_i) + \alpha + \sum_{m\in C(t^e)/\{t^j\}}nmk(m, z^j_i)}{nm(t^e) + K * \alpha + \sum_{m\in C(t^e)/\{t^j\}}nm(m)} 
\end{eqnarray}
, where $C(t^e)$ is the set of all words which cite $t^e$ as their translations in last iteration; $C(t^e)/\{t^j\}$ means to exclude $t^j$ in this set. As the product of the probabilities is usually very small, $p(s^j = t^e)$ has different orders of magnitude for each $t^e$, so the sampling of $s^e$ can be approximated by selecting the one with largest probability. We use the following equation instead:
\begin{eqnarray}
& s^j \approx  \\ \nonumber
& \arg \max_{t^e} \sum_i \log\frac{nmk(t^e, z^j_i) + \alpha + \sum_{m\in C(t^e)/\{t^j\}}nmk(m, z^j_i)}{nm(t^e) + K * \alpha + \sum_{m\in C(t^e)/\{t^j\}}nm(m)} 
\end{eqnarray}

After sampling the translation selection $s^j = t^e$ for $t^j$, we update the $C(t^e)$ as well as $C(t^{e\prime})$, where $C(t^{e\prime})$ is the previous selection of $s^j$. Then we use a scheme similar to the one used in \ref{sec:infprob} to get topic distribution $\theta$,

\section{Computing Word Similarities to Obtain Translations}
\label{sec:measure}
Once we get the topic distribution of each word, we can use them to calculate the similarity between words. The simplest way to do this is to regard each topic distribution as a vector representation of a word. We can then calculate the cosine similarity between these vectors.
\begin{equation}
Cosine(\theta_m, \theta_c) = \frac{\sum_{k=1}^{K} \theta_{mk} \theta_{ck}} {\sqrt{\sum_{k=1}^{K} \theta_{mk}^2} \sqrt{\sum_{k=1}^{K} \theta_{ck}^2}  }
\end{equation}

Another measure is to use the Kullback-Leibler (KL) divergence. KL divergence is a measure of difference between two probability distributions that is widely used in previous topic model-based approaches.
\begin{equation}
D_{KL}(\theta_m || \theta_c) = \sum_{k=1}^{K} \theta_{mk} \log \frac{\theta_{mk}}{\theta_{ck}}
\end{equation}
Neither cosine similarity nor KL divergence considers the correlation between topics. For a topic model, as we know the topic distribution of each word in addition to knowing the topic itself, we can take advantage of the topic structures by directly modeling the probability of $p(w^e|w^j)$ as the similarity between words $w^e$ and $w^j$. This tells us how likely it is we can generate $w^e$ from $w^j$. We call this similarity measure selProb (selection probability).
\begin{eqnarray*}
&&selProb = p(w^e|w^j) \propto p(w^j|\theta_{w^e})\\
&&= \prod_{i = 1}^{n} \sum_{z^j = 1}^{K} p(d^j_i|z^j,\phi^j) p(z^j|\theta_{w^e})
\end{eqnarray*}
Then, we can select the most similar word in the target language as the translation.
\begin{eqnarray*}
&&\arg\max_{w^e} \log p(w^j|\theta_{w^e})\\
&&= \arg\max_{w^e} \sum_{i = 1}^{n} \log{ \sum_{z^j = 1}^{K} p(d^j_i|z^j,\phi^j) p(z^j|\theta_{w^e})}
\end{eqnarray*}

\section{Experimental Results}
\label{sec:experiment}

\subsection{Experiment Data}
We use two Japanese-English domain-specific corpora for our experiments. The first corpus comes from a bilingual law dataset \footnote{http://www.phontron.com/jaen-law/index-ja.html}. We selected it for our experiments because it has an associated law dictionary that can be directly used as test data. Although this corpus was originally parallel, we do not use it as a parallel data source. We randomly select 150,000 paragraphs for each language (each paragraph is seen as one document) and the paragraphs are not kept aligned. The other dataset is a collection of car complaints from MLIT\footnote{http://www.mlit.go.jp/jidosha/carinf/rcl/defects.html} and NHTSA\footnote{http://www-odi.nhtsa.dot.gov/downloads/index.cfm}. The data, which is unbalanced, includes 351,811 short English documents and 32,059 short Japanese documents.

\begin{table*}[ht]
\caption{Accuracies of translations on law data and car data.}
\label{tb:acc}
\centering
\begin{tabular}{|c||c|c||c|c||c|c|}
\hline
& \multicolumn{4}{c|}{Law} & \multicolumn{2}{c|}{Car}\\
\hline
 & Acc1\ & Acc10 & Acc1\_new & Acc10\_new & Acc1 & Acc10\\
\hline
TFIDF & 57.6\% & 72.1\% & 5.2\% & 20.3\% &1.3\% & 3.3\% \\
\hline
LP & 44.2\% & 71.7\% & 3.8\% & 15.0\% & 2.7\% & 7.3\% \\
\hline
Mixed word embedding & 43.8\% & 71.9\% & 1.5\% & 15.0\% & 4.7\% & 8.7\% \\
\hline
BiLDA + cosine & 56.3\% & 73.4\% & 4.5\% & 9.8\% & 4\% & 12\% \\
\hline
BiLDA + KLD & 58.2\% & 74.8\% & 6.8\% & 18.0\% & 4.7\% & 9.3\% \\
\hline
BiLDA + selProb & 53.3\% & 74.9\% & 6.0\% & 17.3\% & 5.3\% & 12.7\% \\
\hline
BiLDA\_all + cosine & 56.3\% & 72.6\% & 1.5\% & 9.0\% &4.7\% & 9.3\% \\
\hline
BiLDA\_all + KLD & 56.4\% & 72.7\% & 3.0\% & 10.5\% & 4\% & 6.7\% \\
\hline
BiLDA\_all + selProb & 52.1\% & 73.8\% & 2.25\% & 12.0\% & 6\% & 13.3\% \\
\hline
ProbBiLDA + cosine & 55.7\% & 74.7\% & 9.8\% & 17.3\% & 2.7\% & 7.3\% \\
\hline
ProbBiLDA + KLD & 54.8\% & 75.2\% & 5.3\% & 18.0\% &2.7& 4.0\% \\
\hline
ProbBiLDA + selProb & 57.4\% & 77.4\% & 11.3\% & 24.1\% & 6\% & 11.3\% \\
\hline
BlockProbBiLDA + cosine& 58.1\% & 76.1\% & 9.0\% & 23.3\% & 6.7\% & 10\% \\
\hline
BlockProbBiLDA + KLD & 59.3\% & 77.6\% & \bf{14.3}\% & 29.3\% & 4.0\% & 8.0\% \\
\hline
BlockProbBiLDA + selProb & \bf{60.5}\% & \bf{78.1}\% & \bf{14.3}\% & \bf{31.6\%} & \bf{8.0}\% & \bf{14.6}\% \\
\hline
\end{tabular}

\end{table*}

The Japanese texts are processed by our own NLP tool to obtain the segmentation and the English texts are tokenized and lemmatized by NLTK\footnote{http://www.nltk.org}. We removed stop words for both Japanese and English texts. We also removed documents containing less than five words. For both corpora, we use the generic Japanese-English dictionary Edict\footnote{http://www.edrdg.org/jmdict/edict.html} and exclude any words that do not appear in our corpora. For the law data, we use the associated law dictionary as the test data. After erasing any words not in the corpora, we are left with 840 words for test data, within which 133 words are not in the generic dictionary and 254 do not have correct translations in the generic dictionary. In terms of the car data, we do not have any technical dictionary for cars, so we manually annotate 150 technical words not covered by the seed dictionary. In contrast to \cite{andrade2010robust} and \cite{tamura2012bilingual}, we test not only nouns but also words of other parts of speech.

\subsection{Comparisons with Other Models}
We compare our approaches with several previous approaches. 
\begin{itemize}
\item {\bf TFIDF} \cite{fung1998ir} is a classical lexicon extraction method. It uses tf-idf weights for contextual seed words to obtain the context vectors and then uses cosines similarity to rank candidate translations. 
\item {\bf Label propagation (LP)} propagates seed distributions on a graph representing relations among words, and translation pairs are extracted by identifying word pairs with a high similarity in the seed distributions. It is claimed that it could resolve the sparseness problem of context vectors and better utilize the information of seed dictionary. In our experiments, we use the same parameter setting as \cite{tamura2012bilingual} (window size = 4).
\item {\bf Mixed word embedding} \cite{gouws2015simple} is a simple but efficient cross-lingual word embedding that can handle the problem of multiple translation (unlike the one-to-one seed dictionary used in \cite{mikolov2013exploiting}). After obtaining the word embeddings, we use cosine similarity to rank the candidates. 
\end{itemize}

We implemented two bilingual LDA \ref{sec:BiLDA} models as our baselines and compare them with the ProbBiLDA and BlockProbBiLDA. For all the models, we set the hyperparameters as follows: $\alpha=\alpha_{\phi} = 0.5$, $\beta =0.01$, and the topic number is 50.   We use three similarity measures to rank the candidate translations: cosine similarity, KL divergence, and selProb. We use Accuracy@K (K=1, 10) as the evaluation metric. Note that for law data, we test not only the overall accuracy of all law dictionary words (Acc1\_full, Acc10\_full) but also the accuracy w.r.t only new words (not in the generic dictionary) (Acc1\_new, Acc10\_new). As for car data, all test words are new, so we do not discriminate them. The results are listed in Table \ref{tb:acc}.

First let us compare the three similarity measures. It is obvious that selProb performs better than the other two in most cases. This supports our assumption that topics are correlated and selProb can better catch the word similarity. In the following comparison, we use only this measure for all topic models.

By comparing LP and TFIDF, we find, somewhat surprisingly, that LP performs worse than TFIDF on law data. This is probably because the seed dictionary is very noisy for the law domain. Because of the noise, LP does not benefit from propagating the seed dictionary—on the contrary, it is badly impacted because error translations are propagated. We may achieve a similar conclusion by comparing BiLDA and BiLDA\_all. However, on car data, the comparison result is reversed, so we can only assume that the seed dictionary for the car domain is cleaner. 

Now let us turn to our two probabilistically linked bilingual topic models, the ProbBiLDA and BlockProbBiLDA. Regardless of the datasets, BlockProbBiLDA is consistently the best model and the ProbBiLDA is the runner-up or at least comparable to other best models. This demonstrates the effectiveness of modeling translation probability. Among the two models, BlockProbBiLDA is better on both datasets. A possible explanation is that BlockProbBiLDA tends to exclude the noise when a word has several translations, while ProbBiLDA is a mixed average of all translations including noise; and BlockProbBiLDA enables words better correlated to their translations than ProbBiLDA.

\section{Conclusion}
\label{sec:conclusion}
In this paper, we proposed a new framework for extracting translations from non-parallel corpora. First we constructed pseudo documents using inverted indexing; and then we introduced two new bilingual topic models, ProbBiLDA and BlockProbBiLDA, to obtain topic distributions for each word. These models are extensions of the classical Bilingual LDA featuring a new hierarchy to integrate the translation probability for multiple translations in the seed dictionary. We advanced the generation of probability to measure the similarity between one candidate word and a given target word. Experimental results show that Bilingual Topic models as well as inverted representation can be effectively utilized for lexicon extraction and that modeling the probabilities of translations in the seed dictionary is helpful as well. 

%% The file named.bst is a bibliography style file for BibTeX 0.99c
\bibliographystyle{named}
\bibliography{ijcai17}

\begin{thebibliography}{}

\bibitem[\protect\citeauthoryear{Andrade \bgroup \em et al.\egroup
  }{2010}]{andrade2010robust}
Daniel Andrade, Tetsuya Nasukawa, and Jun'ichi Tsujii.
\newblock Robust measurement and comparison of context similarity for finding
  translation pairs.
\newblock In {\em Proceedings of the 23rd International Conference on
  Computational Linguistics}, pages 19--27. Association for Computational
  Linguistics, 2010.

\bibitem[\protect\citeauthoryear{Blei \bgroup \em et al.\egroup
  }{2003}]{blei2003latent}
David~M Blei, Andrew~Y Ng, and Michael~I Jordan.
\newblock Latent dirichlet allocation.
\newblock {\em Journal of machine Learning research}, 3(Jan):993--1022, 2003.

\bibitem[\protect\citeauthoryear{Bollegala \bgroup \em et al.\egroup
  }{2015}]{bollegala2015cross}
Danushka Bollegala, Georgios Kontonatsios, and Sophia Ananiadou.
\newblock A cross-lingual similarity measure for detecting biomedical term
  translations.
\newblock {\em PloS one}, 10(6):e0126196, 2015.

\bibitem[\protect\citeauthoryear{Boyd-Graber and
  Blei}{2009}]{boyd2009multilingual}
Jordan Boyd-Graber and David~M Blei.
\newblock Multilingual topic models for unaligned text.
\newblock In {\em Proceedings of the Twenty-Fifth Conference on Uncertainty in
  Artificial Intelligence}, pages 75--82. AUAI Press, 2009.

\bibitem[\protect\citeauthoryear{Chang and Blei}{2009}]{chang2009relational}
Jonathan Chang and Daokvid~M Blei.
\newblock Relational topic models for document networks.
\newblock In {\em International conference on artificial intelligence and
  statistics}, pages 81--88, 2009.

\bibitem[\protect\citeauthoryear{Dietz \bgroup \em et al.\egroup
  }{2007}]{dietz2007unsupervised}
Laura Dietz, Steffen Bickel, and Tobias Scheffer.
\newblock Unsupervised prediction of citation influences.
\newblock In {\em Proceedings of the 24th international conference on Machine
  learning}, pages 233--240. ACM, 2007.

\bibitem[\protect\citeauthoryear{Duong \bgroup \em et al.\egroup
  }{2016}]{duong2016learning}
Long Duong, Hiroshi Kanayama, Tengfei Ma, Steven Bird, and Trevor Cohn.
\newblock Learning crosslingual word embeddings without bilingual corpora.
\newblock In {\em Proceedings of the 2016 Conference on Empirical Methods in
  Natural Language Processing}, page 1285–1295, 2016.

\bibitem[\protect\citeauthoryear{Fung and Yee}{1998}]{fung1998ir}
Pascale Fung and Lo~Yuen Yee.
\newblock An ir approach for translating new words from nonparallel, comparable
  texts.
\newblock In {\em Proceedings of the 17th international conference on
  Computational linguistics-Volume 1}, pages 414--420. Association for
  Computational Linguistics, 1998.

\bibitem[\protect\citeauthoryear{Gouws and S{\o}gaard}{2015}]{gouws2015simple}
Stephan Gouws and Anders S{\o}gaard.
\newblock Simple task-specific bilingual word embeddings.
\newblock In {\em Proceedings of NAACL-HLT}, pages 1386--1390, 2015.

\bibitem[\protect\citeauthoryear{Haghighi \bgroup \em et al.\egroup
  }{2008}]{haghighi2008learning}
Aria Haghighi, Percy Liang, Taylor Berg-Kirkpatrick, and Dan Klein.
\newblock Learning bilingual lexicons from monolingual corpora.
\newblock {\em ACL-08: HLT}, page 771, 2008.

\bibitem[\protect\citeauthoryear{Laws \bgroup \em et al.\egroup
  }{2010}]{laws2010linguistically}
Florian Laws, Lukas Michelbacher, Beate Dorow, Christian Scheible, Ulrich Heid,
  and Hinrich Sch{\"u}tze.
\newblock A linguistically grounded graph model for bilingual lexicon
  extraction.
\newblock In {\em Proceedings of the 23rd International Conference on
  Computational Linguistics: Posters}, pages 614--622. Association for
  Computational Linguistics, 2010.

\bibitem[\protect\citeauthoryear{Lopez}{2008}]{lopez2008statistical}
Adam Lopez.
\newblock Statistical machine translation.
\newblock {\em ACM Computing Surveys (CSUR)}, 40(3):8, 2008.

\bibitem[\protect\citeauthoryear{Mikolov \bgroup \em et al.\egroup
  }{2013}]{mikolov2013exploiting}
Tomas Mikolov, Quoc~V Le, and Ilya Sutskever.
\newblock Exploiting similarities among languages for machine translation.
\newblock {\em arXiv preprint arXiv:1309.4168}, 2013.

\bibitem[\protect\citeauthoryear{Mimno \bgroup \em et al.\egroup
  }{2009}]{mimno2009polylingual}
David Mimno, Hanna~M Wallach, Jason Naradowsky, David~A Smith, and Andrew
  McCallum.
\newblock Polylingual topic models.
\newblock In {\em Proceedings of the 2009 Conference on Empirical Methods in
  Natural Language Processing: Volume 2-Volume 2}, pages 880--889. Association
  for Computational Linguistics, 2009.

\bibitem[\protect\citeauthoryear{Ni \bgroup \em et al.\egroup
  }{2009}]{ni2009mining}
Xiaochuan Ni, Jian-Tao Sun, Jian Hu, and Zheng Chen.
\newblock Mining multilingual topics from wikipedia.
\newblock In {\em Proceedings of the 18th international conference on World
  wide web}, pages 1155--1156. ACM, 2009.

\bibitem[\protect\citeauthoryear{Pekar \bgroup \em et al.\egroup
  }{2006}]{pekar2006finding}
Viktor Pekar, Ruslan Mitkov, Dimitar Blagoev, and Andrea Mulloni.
\newblock Finding translations for low-frequency words in comparable corpora.
\newblock {\em Machine Translation}, 20(4):247--266, 2006.

\bibitem[\protect\citeauthoryear{Rapp}{1995}]{rapp1995identifying}
Reinhard Rapp.
\newblock Identifying word translations in non-parallel texts.
\newblock In {\em Proceedings of the 33rd annual meeting on Association for
  Computational Linguistics}, pages 320--322. Association for Computational
  Linguistics, 1995.

\bibitem[\protect\citeauthoryear{Shezaf and
  Rappoport}{2010}]{shezaf2010bilingual}
Daphna Shezaf and Ari Rappoport.
\newblock Bilingual lexicon generation using non-aligned signatures.
\newblock In {\em Proceedings of the 48th annual meeting of the association for
  computational linguistics}, pages 98--107. Association for Computational
  Linguistics, 2010.

\bibitem[\protect\citeauthoryear{S{\o}gaard \bgroup \em et al.\egroup
  }{2015}]{sogaard2015inverted}
Anders S{\o}gaard, {\v{Z}}eljko Agi{\'c}, H{\'e}ctor~Mart{\'\i}nez Alonso,
  Barbara Plank, Bernd Bohnet, and Anders Johannsen.
\newblock Inverted indexing for cross-lingual nlp.
\newblock In {\em The 53rd Annual Meeting of the Association for Computational
  Linguistics and the 7th International Joint Conference of the Asian
  Federation of Natural Language Processing (ACL-IJCNLP 2015)}, 2015.

\bibitem[\protect\citeauthoryear{Tamura \bgroup \em et al.\egroup
  }{2012}]{tamura2012bilingual}
Akihiro Tamura, Taro Watanabe, and Eiichiro Sumita.
\newblock Bilingual lexicon extraction from comparable corpora using label
  propagation.
\newblock In {\em Proceedings of the 2012 Joint Conference on Empirical Methods
  in Natural Language Processing and Computational Natural Language Learning},
  pages 24--36. Association for Computational Linguistics, 2012.

\bibitem[\protect\citeauthoryear{Vulic and Moens}{2013}]{vulic2013cross}
Ivan Vulic and Marie-Francine Moens.
\newblock Cross-lingual semantic similarity of words as the similarity of their
  semantic word responses.
\newblock In {\em Proceedings of the 2013 Conference of the North American
  Chapter of the Association for Computational Linguistics: Human Language
  Technologies (NAACL-HLT 2013)}, pages 106--116. ACL, 2013.

\bibitem[\protect\citeauthoryear{Vuli{\'c} \bgroup \em et al.\egroup
  }{2011}]{vulic2011identifying}
Ivan Vuli{\'c}, Wim De~Smet, and Marie-Francine Moens.
\newblock Identifying word translations from comparable corpora using latent
  topic models.
\newblock In {\em Proceedings of the 49th Annual Meeting of the Association for
  Computational Linguistics: Human Language Technologies: short papers-Volume
  2}, pages 479--484. Association for Computational Linguistics, 2011.

\bibitem[\protect\citeauthoryear{Zhang \bgroup \em et al.\egroup
  }{2017}]{zhang2017bilingual}
Meng Zhang, Haoruo Peng3~Yang Liu, and Huanbo Luan1~Maosong Sun.
\newblock Bilingual lexicon induction from non-parallel data with minimal
  supervision.
\newblock 2017.

\end{thebibliography}

\end{document}